%% file: report.tex
\pdfoutput=1

\documentclass[11pt]{article}

\usepackage{ACL2023}

\usepackage{times}
\usepackage{latexsym}
\usepackage{tipa}
\usepackage{subfiles}
\usepackage{booktabs}                         
\usepackage[raggedrightboxes]{ragged2e}       
\usepackage{subfiles}                         
\usepackage{longtable}                        
\usepackage{amsmath}                          
\usepackage{graphicx}                         
\usepackage{enumitem}                         
\usepackage{xcolor}                           

\usepackage[T1]{fontenc}

\usepackage[utf8]{inputenc}

\usepackage{microtype}

\usepackage{inconsolata}

\title{Developing a Named Entity Recognition Dataset for Tagalog}

\author{Lester James V. Miranda \\
  \texttt{ljvmiranda@gmail.com} \\}

\begin{document}
\newcommand{\tlunified}{\textsc{TLUnified-NER}}

\maketitle
\begin{abstract}
  We present the development of a Named Entity Recognition (NER) dataset for Tagalog.
  This corpus helps fill the resource gap present in Philippine languages today, where NER resources are scarce.
  The texts were obtained from a pretraining corpora containing news reports, and were labeled by native speakers in an iterative fashion.
  The resulting dataset contains $\sim$7.8k documents across three entity types: Person, Organization, and Location.
  The inter-annotator agreement, as measured by Cohen's $\kappa$, is 0.81.
  We also conducted extensive empirical evaluation of state-of-the-art methods across supervised and transfer learning settings.
  Finally, we released the data and processing code publicly to inspire future work on Tagalog NLP.
\end{abstract}

\section{Introduction}

Tagalog (\texttt{tl}) is one of the major languages in the Philippines with over 28 million speakers in the country \cite{Lewis2009EthnologueL}. 
It constitutes the bulk of Filipino, the country's official language, by sharing its lexical items and grammatical structure.
Despite this fact, there are little to no resources for Tagalog \cite{Cruz2021ImprovingLL}, hampering the development of reliable language technologies.

In this paper, we present \tlunified{},\footnote[1]{The dataset is accessible at \url{https://huggingface.co/datasets/ljvmiranda921/tlunified-ner}} a Tagalog dataset for Named Entity Recognition (NER).
The texts were obtained from TLUnified \cite{Cruz2021ImprovingLL}, a pretraining corpora containing news reports and other types of text.
We focused on NER because of its foundational role in several NLP tasks \citep{Sang2003IntroductionTT,Lample2016NeuralAF}, especially in problems that require the extraction of structured information. 
\tlunified{} consists of $\sim$7.8k documents across three entity types (\textit{Person}, \textit{Organization}, \textit{Location}), modeled closely to the CoNLL Shared Tasks \cite{Sang2002IntroductionTT,Sang2003IntroductionTT}.
Three native speakers conducted the annotation process, resulting to an inter-annotator agreement (IAA) score of 0.81.

We hope that \tlunified{} will allow researchers to build better NER classifiers for Tagalog, and thereby inspire future research on Tagalog NLP through the following contributions:

\begin{enumerate}
  \item We curated and annotated texts from a large pretraining corpora to represent the modern usage of Tagalog in the news domain.
  \item We provided performance baselines across a variety of supervised and transfer learning settings.
\end{enumerate}

\section{Related Work}

\subfile{tables/entity_types}

\paragraph{Tagalog language} 
Tagalog is an agglutinative language within the Austronesian family \cite{Kroeger1992PhraseSA}. 
It uses the Latin script for its writing system with 28 letters in its alphabet. 
Twenty-six letters are the same as in English, with the addition of \~{N}/\~{n} and Ng/ng.
Tagalog typically follows the VSO word order, but VOS and SVO are also accepted \citep{Schachter1973TagalogRG}.
Although Filipino is the country's official language, it has little to no linguistic differences with Tagalog.

\paragraph{Tagalog NER datasets}
Unfortunately, resources for Tagalog NER are meager.
One major resource is WikiANN \cite{Pan2017CrosslingualNT}, a silver-standard corpora based on a framework designed for 282 other languages.  
However, the Tagalog portion of WikiANN is full of annotation errors, often misconstruing one entity type as another.
Another NER dataset is the Filipino Storytelling corpora \cite{Costiniano2022CustomCG}. 
Although gold-standard, its entity labels (e.g., \textit{Humans \& Body}, \textit{Natural Environment}, etc.) are too domain-specific for general use.
Finally, the LORELEI project also provides language packs for Tagalog \cite{Strassel2016LORELEILP}, but they're not publicly-accessible.

\tlunified{} aims to fill this resource gap by providing a publicly-assessible gold standard resource for Tagalog NER.

\section{Dataset Collection}

The texts were obtained from \citet{Cruz2021ImprovingLL}'s TLUnified pretraining corpora.
It combines news reports \citep{Cruz2020ExploitingNA}, a preprocessed version of CommonCrawl \citep{OrtizSuarez2019AsynchronousPF}, and several other datasets.
We manually filtered this dataset to contain news reports so as to resemble the CoNLL Shared Tasks \cite{Sang2002IntroductionTT,Sang2003IntroductionTT}.

The texts are diverse. 
It contains articles from different news sites online that ran a published print media or news channel in Metro Manila from 2009 to early 2020.
The topics range from politics, weather, and popular science among others.

\section{Annotation Setup}
\label{sec:annotation_setup}

\begin{figure}[t]
\frame{\includegraphics[width=0.48\textwidth]{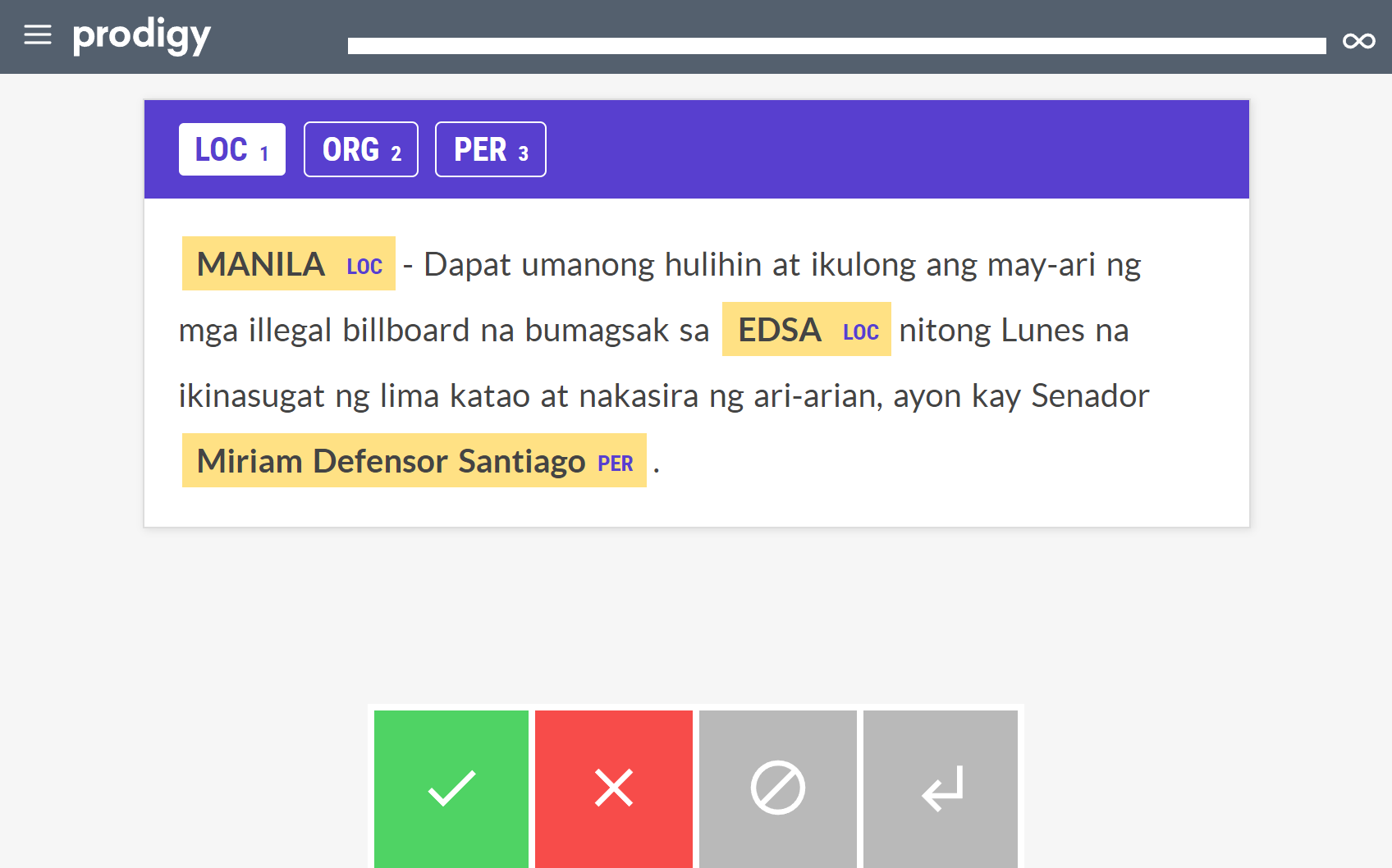}}
\caption{
  Prodigy's annotation interface for a given text. 
  (Translation: \textit{MANILA - The owner of the illegal billboards that fell on EDSA this Monday, injuring five people and damaging property, should be caught and imprisoned according to Senator Miriam Defensor Santiago.})
}
\label{fig:annotation_interface}
\centering
\end{figure}

We used Prodigy as our annotation tool.\footnote[2]{\url{https://prodigy.ai}}
We set up a web server on the Google Cloud Platform and routed the examples through Prodigy's built-in task router.
Figure \ref{fig:annotation_interface} shows the labeling interface as seen by the annotator.
Finally, we used the \texttt{ner.manual} recipe to highlight spans during the annotation process.
We used three entity labels for \tlunified{} as shown in Table \ref{table:entity_types}.
Unlike CoNLL, we decided to exclude the \textit{Miscellaneous (MISC)} tag to reduce confusion.

\subfile{tables/dataset_statistics.tex}

\paragraph{Annotation Process}
The annotation process was done iteratively with three annotators (including the author) who are native Tagalog speakers.
Given a set annotation budget, we paid the annotators above the country's minimum daily wage.
Each annotation round spans for two to three weeks, for a total of six rounds (18 weeks).
The annotators labeled the same batch of examples to ensure high overlap.

After each round, the annotators hold a retrospective meeting and discussed examples they found confusing, inconsistent with the annotation guidelines, and noteworthy.
This process continued until we reached $\sim$10k examples or if we exhausted our annotation budget.
In addition, we also tracked the training curve to determine the quality of the collected annotations.
If the F1-score improved within the last 25\% of the training data, then it is a good sign that obtaining more labels will result to better accuracy.

\paragraph{Annotation Guidelines}
We developed the annotation guidelines in an iterative fashion. 
The Automatic Content Extraction (ACE 2004/05) annotation document \cite{Doddington2004TheAC} heavily inspired our initial draft.
We co-developed the guidelines after each annotation round to improve clarity and reduce disagreements.
These guidelines are accessible on GitHub: \url{https://github.com/ljvmiranda921/calamanCy/tree/master/datasets/tl_calamancy_gold_corpus/guidelines}

\section{Corpus Statistics and Evaluation}

Table \ref{table:dset_stats} shows the final dataset statistics for \tlunified{}.
We also included span- (SD) and boundary-distinctiveness (BD) metrics \cite{Papay2020DissectingSI}.
They measure the KL-divergence of the unigram word distributions between the span (or its boundaries) and the rest of the corpora.
These metrics can be used to gauge the difficulty of the span labeling task, (e.g., more distinct spans means it's ``easier'' to detect them in the text).

\subfile{tables/iaa}

\subsection{Inter-annotator Agreement (IAA)}

Similar to \citet{Brandsen2020CreatingAD}, we measured two types of Cohen's $\kappa$. 
The first metric calculates $\kappa$ for tokens where at least one annotator has made an annotation. 
The second metric computes for all tokens while ignoring the `O' label.
In addition, we had a third measure: the F1-score using one set of annotations as reference \cite{Deleger2012BG}.
We did these computations for each annotator-pair and averaged the results as shown in Table \ref{table:iaa}.

\begin{figure}[t]
\includegraphics[width=0.5\textwidth]{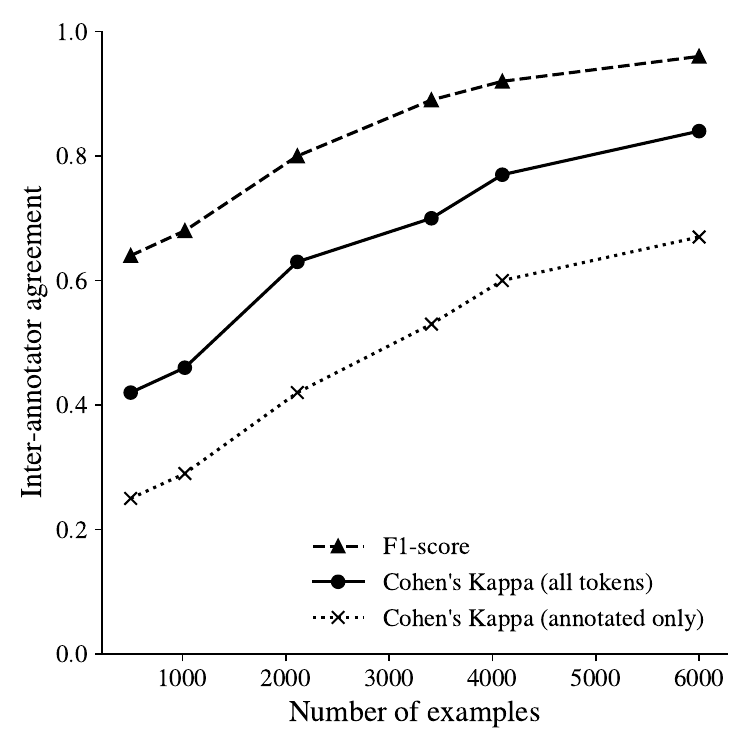}
\caption{Growth of IAA for each annotation round.}
\label{fig:iaa}
\centering
\end{figure}

Finally, Figure \ref{fig:iaa} shows the growth of IAA for each annotation round.
Because of our annotation process, we were able to label the same batch of documents and track the agreement every round.

\subsection{Benchmark results}
\label{subsec:benchmark}

\subfile{tables/results_baseline}

We trained several NER models using spaCy's transition-based parser \cite{Honnibal2020Spacy}.
The state transitions are based on the BILUO sequence encoding scheme and the actions are decided by a convolutional neural network with a maxout \cite{Goodfellow2013MaxoutN} activation function.

While keeping the NER classifier constant, we experimented with various word embeddings that led to the following configurations:

\begin{itemize}
  \item \textbf{Baseline:} we trained the transition-based parser ``from scratch'' without additional information from static or context-sensitive vectors.
  \item \textbf{Static vectors:} we used Tagalog fastText vectors \cite{Bojanowski2016EnrichingWV} and included a simple pretraining process to initialize the weights of the model.
  The pretraining objective asks the model to predict some number of leading and trailing UTF-8 bytes for the words\textemdash a variant of the cloze task.
  \item \textbf{Transformer-based vectors (monolingual):} we used RoBERTa Tagalog \cite{Cruz2021ImprovingLL}, the only pretrained language model for Tagalog, and finetuned it with our annotations.
  \item \textbf{Transformer-based vectors (multilingual):} we tested on XLM-RoBERTa \cite{Conneau2019UnsupervisedCR} and multilingual BERT \cite{Devlin2019BERTPO} for transfer learning.
  These models include Tagalog in their training pool albeit underrepresented.
\end{itemize}

This experimental setup allows us to see the expected performance when training Tagalog NER classifiers using standard techniques.
Table \ref{table:results_baseline} reports the F1-score on the test set across three trials.

\begin{figure}[t]
\includegraphics[width=0.5\textwidth]{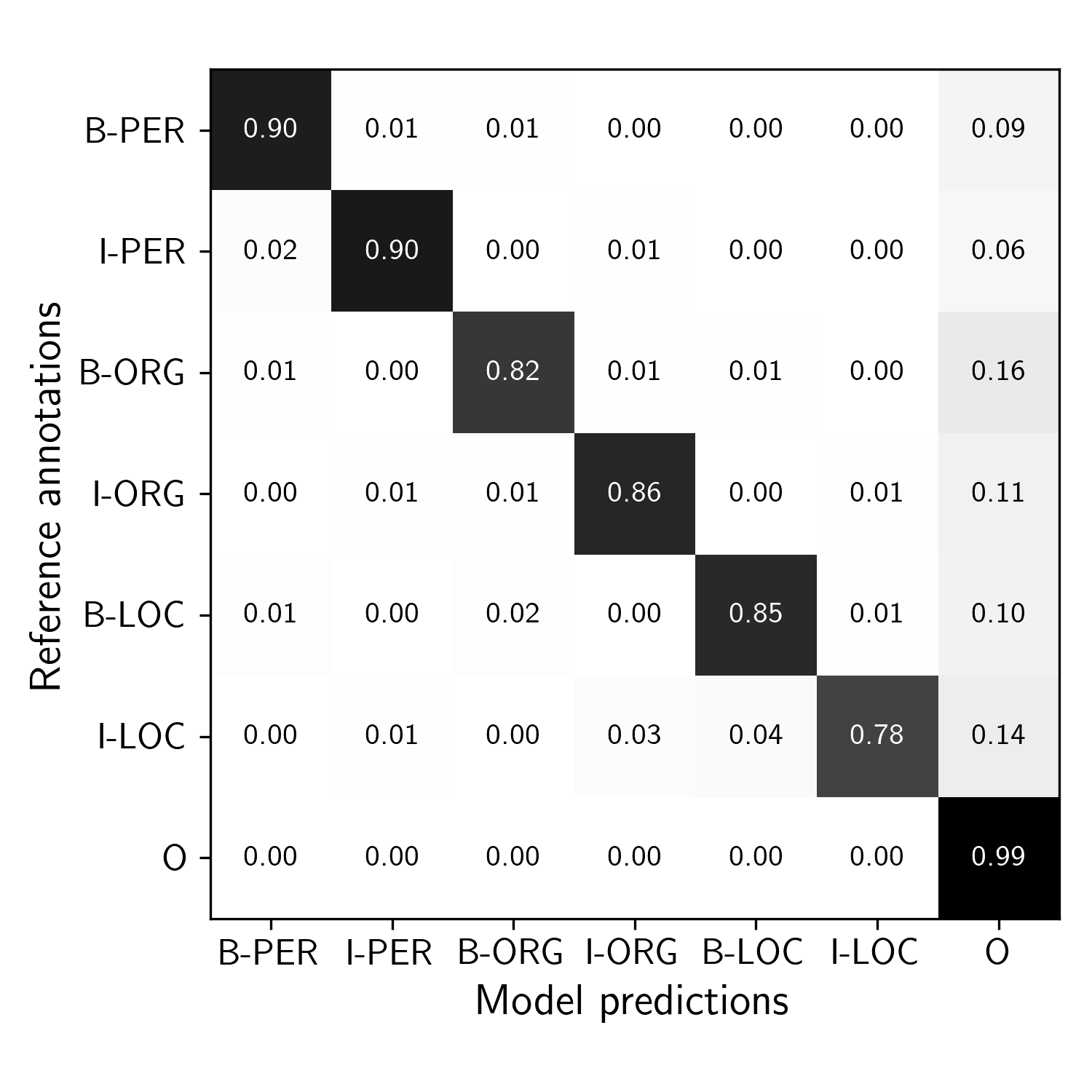}
\caption{Development set confusion matrix of the Baseline model predictions in the IOB format.}
\label{fig:confusion}
\centering
\end{figure}

\subsection{Error analysis}

From our benchmark results, we noticed that most models are having trouble predicting the \textit{Location} or \textit{Organization} tags.
Figure \ref{fig:confusion} shows the confusion matrix of the Baseline model on the development set in the IOB format.

Most of the mistakes came from incorrectly tagging a token with the outside `O' label.
However, we also noticed instances where the model confuses between the lexical and semantic tag of an entity.
For example, in the span, \textit{``\ldots panukala ng Ombudsman\ldots ''} (``\ldots proposed by the Ombudsman\ldots ''), the token \texttt{Ombudsman} might be a Person or Organization depending on the context.
We hypothesize that including context-sensitive training, which the baseline model lacks, can help mitigate this issue.

To test this hypothesis, we experimented on two training configurations.
First, we trained a POS tagger together with our transition-based NER with shared weights.
This process may help provide extra information to the transition-based parser so it can disambiguate between entities.
Second, we finetuned context-sensitive vectors from RoBERTa Tagalog \citep{Cruz2021ImprovingLL} for NER.
Table \ref{table:errors} shows the relative error reduction between LOC and ORG entities.
Given these results, we encourage researchers to utilize context-sensitive vectors such as RoBERTa Tagalog (or other BERT variants) when training models from this corpora.

\subfile{tables/results_errors}

\subfile{tables/results_wikiann}

\subsection{Comparison to WikiANN}

The WikiANN dataset \citep{Pan2017CrosslingualNT} is another resource for Tagalog NER.
However, we found many annotation errors in the dataset, from misclassifications to fragmented sentences.
We investigated how \tlunified{} fares against WikiANN's silver-standard annotations.

We finetuned several models similar to Section \ref{subsec:benchmark} on the Tagalog portion of WikiANN's training set and tested it on \tlunified{}'s test set (and vice-versa).
In order to properly evaluate on WikiANN, we reannotated the test dataset using the same annotation guidelines described in Section \ref{sec:annotation_setup}.

\subfile{tables/results_gold}

Our results in Table \ref{table:results_wikiann} suggest that models built from the \tlunified{} corpus are more performant than with WikiANN.
Additionally, the gap between WikiANN's silver-standard annotations and our corrections is large, as shown in Table \ref{table:results_gold}.
We then posit that the gold-standard nature of \tlunified{} led to better performance than WikiANN, which predominantly consists of text fragments and low-quality annotations.

\subsection{Experiments on large language models}

Large language models (LLMs) have been shown to exhibit multilingual capabilities\textemdash incidental or not \citep{Briakou2023SearchingFN}.
We investigated this property by performing a zero-shot prompting approach on \tlunified{}'s test set across a variety of commercial and open-source LLMs.
Table \ref{table:results_llm} reports the F1-score across three trials.

\subfile{tables/results_llm}

Our results suggest that supervised learning reliably outperforms zero-shot prompting for \tlunified{} given our prompt (see Appendix \ref{appendix:prompt}).
However, we acknowledge that these results are not a definitive comparison between two methods as prompt engineering is unstable with high variance \citep{Webson2022DoPB,Zhao2021CalibrateBU}.
In the future, we plan to explore different prompting techniques such as PromptNER \citep{Ashok2023PromptNER} and chain-of-thought \citep{Wei2023CoT} to uncover the language models' full capabilities.

\section{Conclusion}

In this paper, we introduced \tlunified{}, a Named Entity Recognition dataset for Tagalog.
Unlike other Tagalog NER datasets, \tlunified{} is publicly-accessible and gold standard.
Our iterative annotation process, together with our inter-annotator agreement, shows that the corpus is of high quality.
In addition, our benchmarking results suggest that the task is learnable even with a simple baseline method.
We hope that \tlunified{} fills the resource gap present in Tagalog NLP today.
In the future, we plan to create a more fine-grained (and perhaps, overlapping) NER tag set similar to the ACE project and expand on other major Philippine languages.
Finally, the dataset is available online (\url{https://huggingface.co/datasets/ljvmiranda921/tlunified-ner}) and we encourage researchers to improve upon our benchmark results.

\section*{Limitations}

The \tlunified{} corpora is comprised mostly by news reports.
Although the texts demonstrate the standard usage of Tagalog, its domain is limited.
In addition, we only trained a transition-based parser model for our NER classifier.
In the future, we plan to extend these benchmarks and include CRFs or other tools such as Stanford Stanza.

\section*{Acknowledgements}

We would like to express our gratitute to all those who contributed to the completion of this resource.
We extend our appreciation to the anonymous reviewers for their constructive comments, which greatly improved the quality of this paper. 

\bibliography{custom}
\bibliographystyle{acl_natbib}

\appendix

\section{Appendix}

\subsection{Zero-shot prompt template}
\label{appendix:prompt}

{
\itshape
You are an expert Named Entity Recognition 
(NER) system. Your task is to accept Text as 
input and extract named entities for the set 
of predefined entity labels. From the Text 
input provided, extract named entities for 
each label in the following format:

\begin{itemize}[noitemsep,topsep=0pt]
\item PER: <comma delimited list of strings>
\item ORG: <comma delimited list of strings>
\item LOC: <comma delimited list of strings>
\end{itemize}

Below are definitions of each label to help 
aid you in what kinds of named entities to 
extract for each label. Assume these 
definitions are written by an expert and 
follow them closely.

\begin{itemize}[noitemsep,topsep=0pt]
\item PER: PERSON
\item ORG: ORGANIZATION
\item LOC: LOCATION OR GEOPOLITICAL ENTITY
\end{itemize}

Text: \{\{~text~\}\}
}

\subsection{Reproducibility}

All the experiments and models in this paper are available publicly. 
Readers can head over to \url{https://github.com/ljvmiranda921/calamanCy/reports/aacl} for all related code and assets.
Note that the XLM-RoBERTa and multilingual BERT experiments may at least require a T4 or V100 GPU.

\end{document}

%% file: tables/entity_types.tex
\begin{table*}[t]
\begin{tabular}{@{}p{2cm}p{9cm}p{4cm}@{}}
\toprule
Entity             & Short Description                                                                                                              & Examples                                       \\ \midrule
Person (PER)       & Person entities limited to humans. It may be a single individual or group.                                                     & Juan de la Cruz, Jose Rizal, Quijano de Manila \\
Organization (ORG) & Organization entities limited to corporations, agencies, and other groups of people defined by an organizational structure.    & Meralco, DPWH, United Nations                  \\
Location (LOC)     & Location entities are geographical regions, areas, and landmasses. Geo-political entities are also included within this group. & Pilipinas, Manila, CALABARZON, Ilog Pasig        \\ \bottomrule
\end{tabular}
\caption{
Entity types used for annotating \tlunified{} (derived from the TLUnified pretraining corpus of \citealp{Cruz2021ImprovingLL}).
}
\label{table:entity_types}
\end{table*}

%% file: tables/dataset_statistics.tex
\begin{table*}[t]
\centering
\begin{tabular}{@{}lrrrrrrrr@{}}
\toprule
Dataset     & Examples & Tokens &  PER  & ORG  & LOC  &  Length  & SD    & BD       \\ \midrule
Training    & 6252     & 198579 &  6418 & 3121 & 3296 &  1.49    & 2.66  & 1.26     \\
Development & 782      & 25069  &  793  & 392  & 409  &  1.51    & 2.77  & 1.37     \\
Test        & 782      & 25100  &  818  & 423  & 438  &  1.48    & 2.77  & 1.34     \\ \bottomrule
\end{tabular}
\caption{
    Dataset statistics for \tlunified{}.
    It shows the number of examples, number of tokens, and span-level statistics.
    SD stands for span distinctiveness whereas BD is boundary distinctiveness \cite{Papay2020DissectingSI}.
}
\label{table:dset_stats}
\end{table*}

%% file: tables/iaa.tex
\begin{table}[t]
\begin{tabular}{@{}p{6cm}p{1.25cm}@{}}
\toprule
Metric &  IAA  \\ \midrule
Cohen's $\kappa$ on all tokens & 0.81 \\ 
Cohen's $\kappa$ on annotated tokens only & 0.65 \\
F1 score & 0.91 \\ \bottomrule
\end{tabular}
\caption{
    Inter-annotator agreement (IAA) measurements.
    We obtained these values by computing for the pairwise comparisons on all annotator-pairs and averaging the results.
}
\label{table:iaa}
\end{table}

%% file: tables/results_baseline.tex
\begin{table*}[t]
\centering
\begin{tabular}{@{}lrrrr@{}}
\toprule
Word Embeddings & Person & Organization &  Location  & Overall       \\ \midrule
Baseline (no additional embeddings)          & 87.85$\pm$0.01 & 74.80$\pm$0.02 & 81.03$\pm$0.01 & 84.57$\pm$0.02 \\ 
fastText  \cite{Bojanowski2016EnrichingWV}   & 91.20$\pm$0.02 & 85.39$\pm$0.03 & 88.38$\pm$0.01 & 88.90$\pm$0.01 \\ 
RoBERTa Tagalog \cite{Cruz2021ImprovingLL}   & 92.18$\pm$0.01 & 87.30$\pm$0.00 & 90.01$\pm$0.02 & 90.34$\pm$0.02 \\
XLM-RoBERTa \cite{Conneau2019UnsupervisedCR} & 91.95$\pm$0.04 & 84.84$\pm$0.02 & 88.92$\pm$0.01 & 88.03$\pm$0.03\\
Multilingual BERT \cite{Devlin2019BERTPO}    & 90.78$\pm$0.03 & 85.08$\pm$0.01 & 88.45$\pm$0.03 & 87.40$\pm$0.02 \\
\bottomrule
\end{tabular}
\caption{
    Benchmark results on \tlunified{} across different word embeddings using spaCy's transition-based parser \cite{Honnibal2020Spacy}.
    Reported results are F1-scores on the test set across three trials.
}
\label{table:results_baseline}
\end{table*}

%% file: tables/results_errors.tex
\begin{table}[t]
\centering
{
\begin{tabular}{@{}lrr@{}}
\toprule
                         & \multicolumn{2}{c}{Rel. error reduction} \\ \cmidrule{2-3}
Embeddings set-up        & ORG        & LOC      \\ \midrule
Shared                   & +5\%       & +3\%     \\ 
Context-sensitive        & +12\%      & +18\%    \\
\bottomrule
\end{tabular}
}
\caption{
    Relative error reduction (with respect to the Baseline) for classifying ORG and LOC entities.
    Reported results are F1-scores on the development set.
}
\label{table:errors}
\end{table}

%% file: tables/results_wikiann.tex
\begin{table*}[t]
\centering
{
\begin{tabular}{@{}lrr@{}}
\toprule
           & \multicolumn{2}{c}{Training dataset} \\ \cmidrule{2-3}
Model      & WikiANN   & \tlunified{}             \\ \midrule
Baseline (no additional embeddings)          & 19.92$\pm$0.03 & \textbf{30.24$\pm$0.02}  \\ 
fastText  \cite{Bojanowski2016EnrichingWV}   & 24.41$\pm$0.01 & \textbf{45.09$\pm$0.02}  \\ 
RoBERTa Tagalog \cite{Cruz2021ImprovingLL}   & 23.38$\pm$0.02 & \textbf{58.90$\pm$0.03}  \\
XLM-RoBERTa \cite{Conneau2019UnsupervisedCR} & 31.28$\pm$0.01 & \textbf{57.67$\pm$0.01}  \\
Multilingual BERT \cite{Devlin2019BERTPO}    & 29.20$\pm$0.03 & \textbf{59.26$\pm$0.03}  \\
\bottomrule
\end{tabular}

}

\caption{
    Cross-dataset comparison between WikiANN \citep{Pan2017CrosslingualNT} and \tlunified{}. 
    We trained a model from WikiANN then applied it to \tlunified{} (and vice-versa).
    Reported results are F1-scores on the test set across three trials.
}
\label{table:results_wikiann}
\end{table*}

%% file: tables/results_gold.tex
\begin{table}[t]
\centering
{
\begin{tabular}{@{}lr@{}}
\toprule
Entity label       & F1-score  \\ \midrule
Person (PER)       & 67.95     \\ 
Organization (ORG) & 00.59     \\ 
Location (LOC)     & 35.17     \\
\bottomrule
\end{tabular}

}
\caption{
    Comparing the overlap between the original (silver-standard) WikiANN annotations against our reannotated version.
}
\label{table:results_gold}
\end{table}

%% file: tables/results_llm.tex
\begin{table}[t]
\centering
\begin{tabular}{@{}lr@{}}
\toprule
Model & F1-score \\ \midrule
GPT-4 \citep{OpenAI2023GPT4}                & 65.89$\pm$0.44 \\ 
GPT-3.5-turbo                               & 53.05$\pm$0.42 \\ 
Claude v1 \citep{Anthropic2023Claude}       & 58.88$\pm$0.03 \\
Command \citep{Command2023Cohere}           & 25.48$\pm$0.11 \\
Dolly v2* \citep{DatabricksBlog2023DollyV2} & 13.07$\pm$0.14 \\
Falcon* \citep{Falcon40b}                   & 8.65$\pm$0.04  \\
StableLM v2* \citep{StableLM2023}           & 0.25$\pm$0.03  \\
OpenLLaMa* \citep{openlm2023openllama}      & 15.09$\pm$0.48 \\
\bottomrule
\end{tabular}
\caption{
    Benchmark results on \tlunified{} across a variety of open-source and commercial LLMs.
    We used the 7B-parameter variants for models denoted with an asterisk ($\ast$) due to budget constraints.
}
\label{table:results_llm}
\end{table}